\documentclass[runningheads]{llncs}
\usepackage{graphicx}
\usepackage{comment}
\usepackage{amsmath,amssymb} 
\usepackage{amsfonts}
\usepackage{color}
\usepackage{times}
\usepackage{epsfig}
\usepackage{helvet}
\usepackage{courier}
\usepackage{url}
\usepackage{rotating}
\usepackage{xspace}
\usepackage{booktabs}
\usepackage{nccmath}
\usepackage{nicefrac}
\usepackage{microtype}
\usepackage{adjustbox}		
\usepackage{tabu}
\usepackage{calc}			
\usepackage{wrapfig}
\usepackage{tabularx}
\usepackage{makecell}
\usepackage{multirow}
\usepackage{float}
\usepackage[hang,flushmargin]{footmisc}
\usepackage[neverdecrease]{paralist}
\usepackage{caption}
\usepackage{subcaption}
\usepackage{algorithm}
\usepackage{algorithmic}
\usepackage[table]{xcolor}
\usepackage{colortbl}
\usepackage{pifont}
\usepackage[pagebackref=false,breaklinks=true,letterpaper=true,colorlinks,citecolor=blue,linkcolor=blue,bookmarks=false]{hyperref}

\usepackage[width=122mm,left=12mm,paperwidth=146mm,height=193mm,top=12mm,paperheight=217mm]{geometry}

\newcommand{\calF}{{\mathcal{F}}}

\newcommand{\calL}{{\mathcal{L}}}

\newcommand{\be}{\begin{eqnarray}}
\newcommand{\ee}{\end{eqnarray}}
\newcommand{\bee}{\begin{eqnarray*}}
\newcommand{\eee}{\end{eqnarray*}}

\newcommand{\matrixb}{\left[ \begin{array}}
\newcommand{\matrixe}{\end{array} \right]}

\definecolor{CuGray}{gray}{0.9}
\newcolumntype{g}{>{\columncolor{CuGray}}c}

\newcommand{\cmark}{\ding{51}}%
\newcommand{\xmark}{\ding{55}}%

\makeatletter
\DeclareRobustCommand\onedot{\futurelet\@let@token\@onedot}
\def\@onedot{\ifx\@let@token.\else.\null\fi\xspace}

\newcommand{\dashrule}[1][black]{%
  \color{#1}\rule[\dimexpr.5ex-.2pt]{4pt}{.4pt}\xleaders\hbox{\rule{4pt}{0pt}\rule[\dimexpr.5ex-.2pt]{4pt}{.4pt}}\hfill\kern0pt%
}

\newcommand*{\Scale}[2][4]{\scalebox{#1}{$#2$}}%

\newcommand{\figref}[1]{Fig.~\ref{#1}}
\newcommand{\tabref}[1]{Table~\ref{#1}}
\newcommand{\eqnref}[1]{Eq.~(\ref{#1})}
\newcommand{\secref}[1]{Sec.~\ref{#1}}

\def\eg{\emph{e.g}\onedot}

\def\etal{\emph{et al}\onedot}

\begin{document}
\pagestyle{headings}
\mainmatter
\def\ECCVSubNumber{354}  

\title{Instance-wise Depth and Motion Learning from Monocular Videos} 

\titlerunning{Instance-wise Depth and Motion Learning from Monocular Videos}
%
\author{
    \hspace{0mm}Seokju Lee$^1$
    \hspace{6mm}Sunghoon Im$^2$
    \hspace{6mm}Stephen Lin$^3$
    \hspace{6mm}In So Kweon$^1$
}
\authorrunning{S. Lee \etal}
%
\institute{
\hspace{0mm}$^1$KAIST
\hspace{6mm}$^2$DGIST
\hspace{6mm}$^3$Microsoft Research
\vspace{1mm}
\\
{\tt\small seokju91@gmail.com}
}
\maketitle

\begin{abstract}
We present an end-to-end joint training framework that explicitly models 6-DoF motion of multiple dynamic objects, ego-motion and depth in a monocular camera setup without supervision.
Our technical contributions are three-fold.
First, we propose a differentiable forward rigid projection module that plays a key role in our instance-wise depth and motion learning.
Second, we design an instance-wise photometric and geometric consistency loss that effectively decomposes background and moving object regions.
Lastly, we introduce a new auto-annotation scheme to produce video instance segmentation maps that will be utilized as input to our training pipeline.
These proposed elements are validated in a detailed ablation study.
Through extensive experiments conducted on the KITTI dataset, our framework is shown to outperform the state-of-the-art depth and motion estimation methods.
Our code and dataset will be available at \url{https://github.com/SeokjuLee/Insta-DM}.
\end{abstract}


\begin{figure}[t]
\centering
\includegraphics[width=0.95\textwidth]{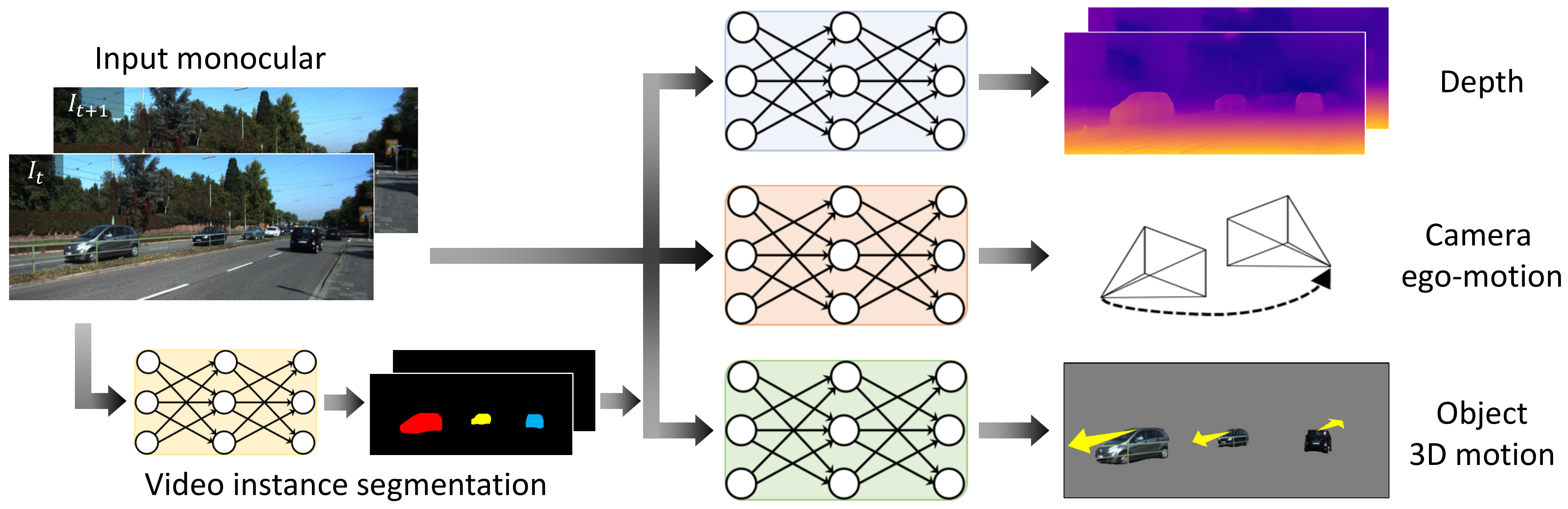}
\vspace{-2mm}
\caption{Overview of our system. The proposed method learns 3D motion of multiple dynamic objects, ego-motion and depth, leveraging an intermediate video instance segmentation. }
\vspace{-4mm}
\label{teaser}
\end{figure}

\section{Introduction}
\vspace{-1mm}

Knowledge of the 3D environment structure and the motion of dynamic objects is essential for autonomous navigation~\cite{geiger20143d,shashua2004pedestrian}.
The 3D structure is valuable because it implicitly models the relative position of the agent, and it is also utilized to improve the performance of high-level scene understanding tasks such as detection and segmentation~\cite{lee2017vpgnet,shin2019roarnet,behley2019iccv,lee2019visuomotor}.
Besides scene structure, the 3D motion of the agent and traffic participants such as pedestrians and vehicles is also required for safe driving.
The relative direction and speed between them are taken as the primary inputs for determining the next direction of travel.

Recent advances in deep neural networks (DNNs) have led to a surge of interest in depth prediction using monocular images~\cite{eigen2014depth,garg2016unsupervised} and stereo images~\cite{mayer2016large,chang2018pyramid}, as well as in optical flow estimation~\cite{dosovitskiy2015flownet,sun2018pwc,lv2018learning}.
These supervised methods require a large amount and broad variety of training data with ground-truth labels.
Studies have shown significant progress in unsupervised learning of depth and ego-motion from unlabeled image sequences~\cite{zhou2017unsupervised,godard2017unsupervised,wang2018learning,mahjourian2018unsupervised,ranjan2019collaboration}.
The joint optimization framework uses a network for predicting single-view depth and pose, and exploits view synthesis of images in the sequence as the supervisory signal.
However, these works ignore or mask out regions of moving objects for pose and depth inference.

In this work, rather than consider moving object regions as a nuisance~\cite{zhou2017unsupervised,godard2017unsupervised,wang2018learning,mahjourian2018unsupervised,ranjan2019collaboration}, we utilize them as an important clue for estimating 3D object motions.
This problem can be formulated as motion factorization of object motion and ego-motion.
Factorizing object motion in monocular sequences is a challenging problem, especially in complex urban environments that contain many dynamic objects.
Moreover, deformable dynamic objects such as humans make the problem more difficult because of the greater inaccuracy in their correspondence~\cite{russell2014video}.

To address this problem, we propose a novel framework that explicitly models 3D motions of dynamic objects and ego-motion together with scene depth in a monocular camera setting.
Our unsupervised method relies solely on monocular video for training (without any ground-truth labels) and imposes a photo-consistency loss on warped frames from one time step to the next in a sequence.
Given two consecutive frames from a video, the proposed neural network produces depth, 6-DoF motion of each moving object, and the ego-motion between adjacent frames as shown in~\figref{teaser}.
In this process, we leverage the instance mask of each dynamic object, obtained from an off-the-shelf instance segmentation module.

Our main contributions are the following:

\noindent\textbf{Forward image warping}
Differentiable depth-based rendering (which we call inverse warping) was introduced in~\cite{zhou2017unsupervised}, where the target view $I_t$ is reconstructed by sampling pixels from a source view $I_s$ based on the target depth map $D_t$ and the relative pose $T_{t\rightarrow s}$.
The warping procedure is effective in static scene areas, but the regions of moving objects cause warping artifacts because the 3D structure of the source image $I_s$ may become distorted after warping based on the target image's depth $D_t$~\cite{casser2019depth}.
To build a geometrically plausible formulation, we introduce forward warping which maps the source image $I_s$ to the target viewpoint based on the source depth $D_s$ and the relative pose $T_{s\rightarrow t}$.
There is a well-known remaining issue with forward warping that the output image may have holes.
Thus, we propose the differentiable and hole-free forward warping module that works as a key component in our instance-wise depth and motion learning from monocular videos.
The details are described in~\secref{sec:3_2}.

\noindent\textbf{Instance-wise photometric and geometric consistency}
Existing works~\cite{cao2019learning,lee2019learning} have successfully estimated independent object motion with stereo cameras.
Approaches based on stereo video can explicitly separate static and dynamic motion by using stereo offset and temporal information.
On the other hand, estimation from monocular video captured in the dynamic real world, where both agents and objects are moving, suffers from motion ambiguity, as only temporal clues are available.
To address this issue, we introduce instance-wise view synthesis and geometric consistency into the training loss.
We first decompose the image into background and object (potentially moving) regions using a predicted instance mask.
We then warp each component using the estimated single-view depth and camera poses to compute photometric consistency.
We also impose a geometric consistency loss for each instance that constrains the estimated geometry from all input frames to be consistent.
\secref{sec:3_3} presents our technical approach (our loss and network details) for inferring 3D object motion, ego-motion and depth.


\noindent\textbf{Auto-annotation of video instance segmentation}
We introduce an auto-annotation scheme to generate a video instance segmentation dataset, which is expected to contribute to various areas of self-driving research.
The role of this method is similar to that of~\cite{yang2019video}, but we design a new framework that is tailored to driving scenarios on the existing dataset~\cite{geiger2012we,cordts2016cityscapes}.
We modularize this task into instance segmentation~\cite{he2017mask,liu2018path} and optical flow~\cite{sun2018pwc} steps and combine each existing fine-tuned model to generate the tracked instance masks automatically.
Details are described in~\secref{sec:4_2}.

\noindent\textbf{State-of-the-art performance} Our unsupervised monocular depth and pose estimation is validated with a performance evaluation, presented in~\secref{sec:4}, which shows that our jointly learned system outperforms earlier approaches.

Our codes, models, and video instance segmentation dataset will be made publicly available.




\vspace{-2mm}
\section{Related Works}
\vspace{-2mm}

\noindent\textbf{Unsupervised depth and ego-motion learning}
Several works~\cite{zhou2017unsupervised,godard2017unsupervised,wang2018learning,mahjourian2018unsupervised,ranjan2019collaboration} have studied the inference of depth and ego-motion from monocular sequences.
Zhou~\etal~\cite{zhou2017unsupervised} introduce an unsupervised learning framework for depth and ego-motion by maximizing photometric consistency across
monocular video frames during training.
Godard~\etal~\cite{godard2017unsupervised} offer an approach replacing the use of explicit depth data during training with easier-to-obtain binocular stereo footage.
It trains a network by searching for the correspondence in a rectified stereo pair that requires only a one-dimensional search.
Wang~\etal~\cite{wang2018learning} show that Direct Visual Odometry (DVO) can be used to estimate the camera pose between frames and the inverse depth normalization leads to a better local minimum.
Mahjourian~\etal~\cite{mahjourian2018unsupervised} combine 3D geometric constraints using Iterative Closest Point (ICP) with a photometric consistency loss.
Ranjan~\etal~\cite{ranjan2019collaboration} propose a competitive collaboration framework that facilitates the coordinated training of multiple specialized neural networks to solve joint problems.
Recently, there have been two works~\cite{bian2019unsupervised,chen2019self} that achieve state-of-the-art performance on depth and ego-motion estimation using geometric constraints.

\begin{figure*}[t]
\centering
\includegraphics[width=0.99\textwidth]{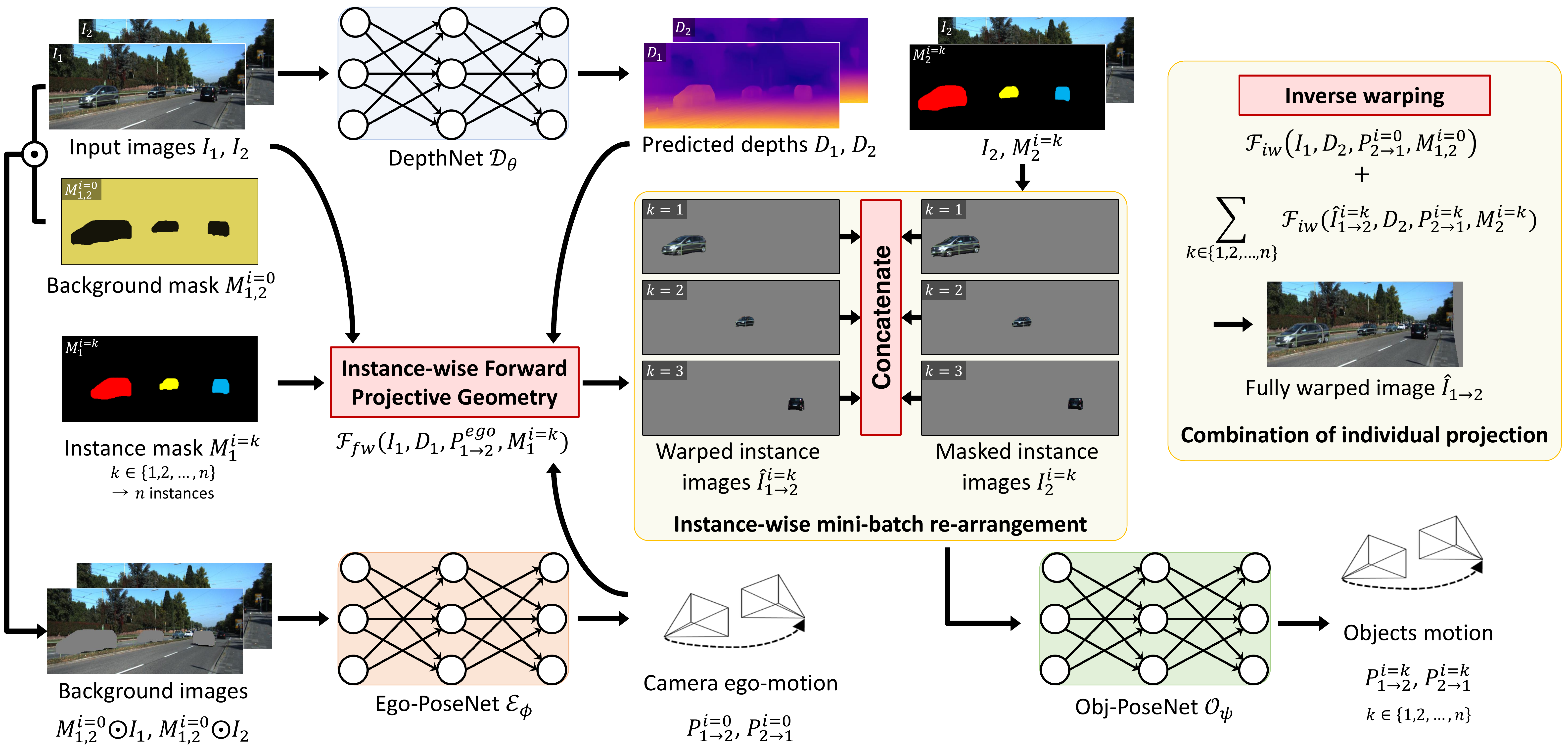}
\vspace{-3mm}
\caption{Overview of the proposed frameworks.}
\vspace{-5mm}
\label{overview}
\end{figure*}

\noindent\textbf{Learning motion of moving objects}
Recently, the joint optimization of dynamic object motion along with depth and ego-motion~\cite{casser2019depth,cao2019learning} has gained interest as a new research topic.
Casser~\etal~\cite{casser2019depth} present an unsupervised image-to-depth framework that models the motion of moving objects and cameras. The main idea is to introduce geometric structure in the learning process, by modeling the scene and the individual objects; camera ego-motion and object motions are learned from monocular videos as input.
Cao~\etal~\cite{cao2019learning} propose a self-supervised framework with a given 2D bounding box to learn scene structure and 3D object motion from stereo videos. They factor the scene representation into independently moving objects with geometric reasoning. However, this work is based on a stereo camera setup and computes the 3D motion vector of each instance using simple mean filtering.

\noindent\textbf{Video instance segmentation}
The task of video instance segmentation (VIS) is to simultaneously conduct detection, segmentation and tracking of instances in videos.
O{\v{s}}ep~\etal~\cite{ovsep2018track} and Voigtlaender~\etal~\cite{voigtlaender2019mots} first extend the box-based tracking task~\cite{ovsep2018track} to pixel-level instance tracking.
Yang~\etal~\cite{yang2019video} generalize the image instance segmentation problem to the video domain.
This task is composed of many sub-tasks: detection, segmentation and tracking. 
Each sub-task is domain-specific and needs end-to-end fine-tuning.


\vspace{-2mm}
\section{Methodology}
\label{sec:3}
\vspace{-3mm}

We introduce an end-to-end joint training framework for instance-wise depth and motion learning from monocular videos without supervision as illustrated in \figref{overview}.
Our main contribution lies in applying the inverse and forward warping in appropriate projection situations.
We propose a geometrically correct warping method in dynamic situations, which is a fundamental problem in 3D geometry.
In this section, we introduce the instance-wise forward projective geometry and the networks for each type of output: DepthNet, Ego-PoseNet, and Obj-PoseNet.
Further, we describe our novel loss functions and how they are designed for back-propagation in decomposing the background and moving object regions.

\vspace{-2mm}
\subsection{Method Overview}
\label{sec:3_1}
\vspace{-2mm}

\noindent\textbf{Baseline}~
Given two consecutive RGB images~$(I_1, I_2)\in \mathbb{R}^{H \times W \times 3}$, sampled from an unlabeled video, we first predict their respective depth maps~$(D_1, D_2)$ via our presented DepthNet~$\mathcal{D}_\theta: \mathbb{R}^{H \times W \times 3} \rightarrow \mathbb{R}^{H \times W}$ with trainable parameters~$\theta$.
By concatenating two sequential images as an input, our proposed Ego-PoseNet~$\mathcal{E}_\phi: \mathbb{R}^{2 \times H \times W \times 3} \rightarrow \mathbb{R}^{6}$, with trainable parameters~$\phi$, estimates the six-dimensional SE(3) relative transformation vectors~$(P_{1 \rightarrow 2}, P_{2 \rightarrow 1})$.
With the predicted depth, relative ego-motion, and a given camera intrinsic matrix~$K \in \mathbb{R}^{3 \times 3}$, we can synthesize an adjacent image in the sequence using an inverse warping operation~$\mathcal{F}_{iw}(I_i, D_j, P_{j \rightarrow i}, K) \rightarrow \hat{I}_{i \rightarrow j}$, where $\hat{I}_{i \rightarrow j}$ is the reconstructed image by warping the reference frame~$I_i$~\cite{zhou2017unsupervised,jaderberg2015spatial}.
As a supervisory signal, an image reconstruction loss, $\mathcal{L}_{rec}=||I_j - \hat{I}_{i \rightarrow j}||_1$, is imposed to optimize the parameters, $\theta$ and $\phi$.

\noindent\textbf{Instance-wise learning}~
The baseline method has a limitation that it cannot handle dynamic scenes containing moving objects.
Our goal is to learn depth and ego-motion, as well as object motions, using monocular videos by constraining them with instance-wise geometric consistencies. We propose an Obj-PoseNet~$\mathcal{O}_\psi: \mathbb{R}^{2 \times H \times W \times 3} \rightarrow \mathbb{R}^{6}$ with trainable parameters~$\psi$, which is specialized to estimate individual object motions.
We annotate a novel video instance segmentation dataset to utilize it as an individual object mask while training the ego-motion and object motions.
The details of the video instance segmentation dataset will be described in Sec.~\ref{sec:4_2}.
Given two consecutive binary instance masks~$(M^{i}_1, M^{i}_2)\in \{0,1\}^{H \times W \times n}$ corresponding to $(I_1, I_2)$, $n$-instances are annotated and matched between the frames.
First, in the case of camera ego-motion, potentially moving objects are masked out and only the background region is fed to Ego-PoseNet.
Secondly, the $n$ binary instance masks are multiplied to the input images and fed to Obj-PoseNet.
For both networks, motions of the $k^{th}$ element are represented as $P^{i=k}_{1 \rightarrow 2}$, where $k=0$ means camera ego-motion from frame $I_1$ to $I_2$.
The details of the motion models will be described in the following subsections.

\noindent\textbf{Training objectives}~
The previous works~\cite{mahjourian2018unsupervised,bian2019unsupervised,chen2019self} imposed geometric constraints between frames, but they are limited to rigid projections.
Regions containing moving objects cannot be constrained with this term and are treated as outlier regions with regard to geometric consistency.
In this paper, we propose instance-wise geometric consistency. We leverage instance masks to impose geometric consistency region-by-region.
Following instance-wise learning, our overall objective function can be defined as follows:
\vspace{-2mm}
\begin{equation}
\Scale[0.99]
{
\begin{aligned}
\calL = \lambda_{r}\calL_{r} + \lambda_{g}\calL_{g} + \lambda_{s}\calL_{s} + \lambda_{t}\calL_{t}  + \lambda_{h}\calL_{h},
\end{aligned}
}
\label{eq_loss}
\vspace{-0mm}
\end{equation}
where
$(\calL_{r}, \calL_{g})$ are the reconstruction and geometric consistency losses applied on each instance region including the background, $\calL_{s}$ stands for the depth smoothness loss, and $(\calL_{t}, \calL_{h})$ are the object translation and height constraint losses.
$\{\lambda_{r}, \lambda_{g}, \lambda_{s}, \lambda_{t}, \lambda_{h}\}$ is the set of loss weights.
We train the models in both forward ($I_1 \rightarrow I_2$) and backward ($I_2 \rightarrow I_1$) directions to maximally use the self-supervisory signals.
In the following subsections, we introduce how to constrain the instance-wise consistencies.

\vspace{-2mm}
\subsection{Forward Projective Geometry}
\label{sec:3_2}
\vspace{-2mm}

\begin{figure}[t]
\centering
\captionsetup[subfigure]{aboveskip=0pt}
\captionsetup[subfigure]{belowskip=0pt}
\begin{tabular}{c@{\hspace{0mm}}c@{\hspace{0mm}}c@{\hspace{0mm}}}
	\subcaptionbox{\label{warping}}{\includegraphics[width=0.32\textwidth]{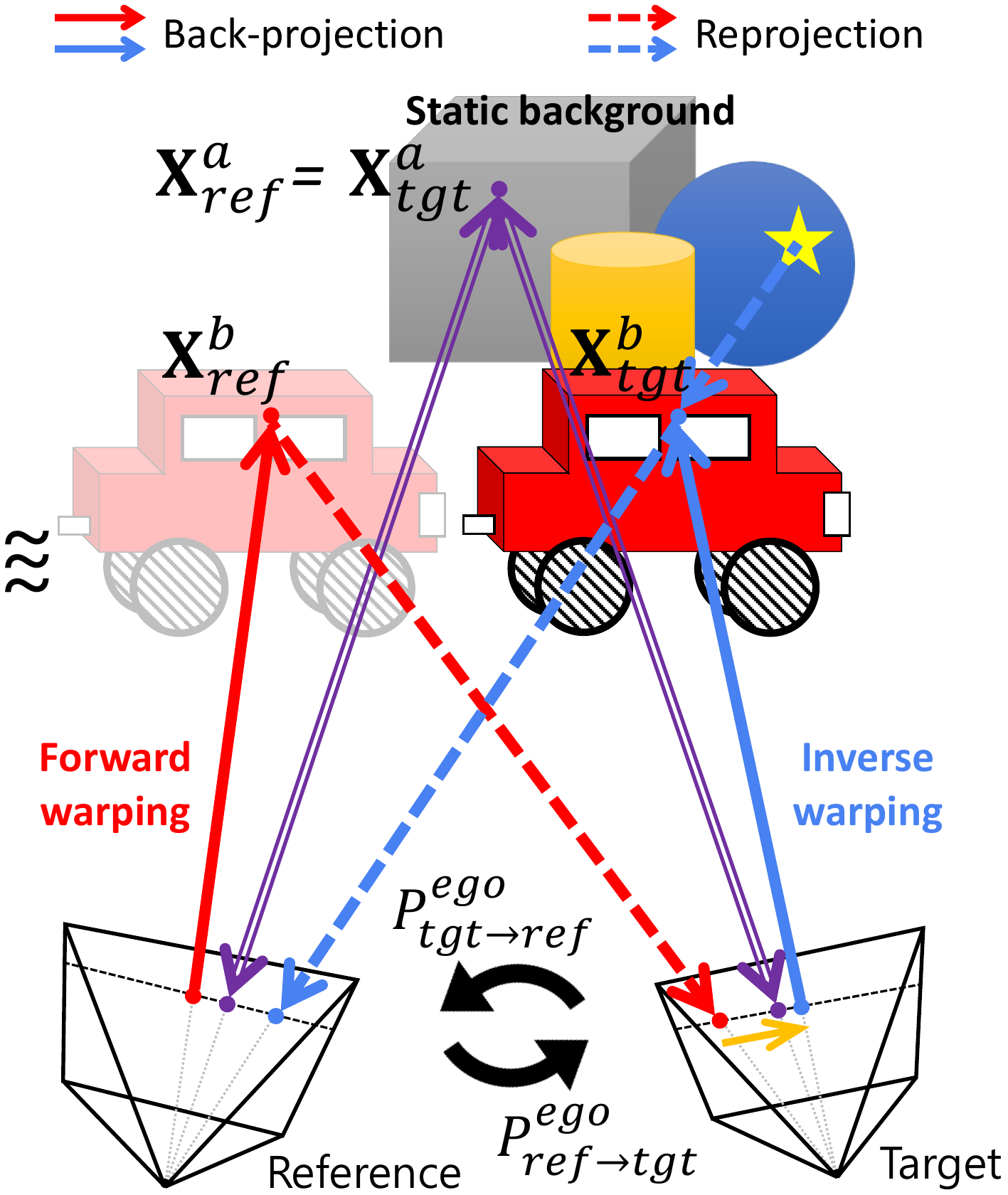}\vspace{+3mm}}
	\hspace{+0.5mm}
	\subcaptionbox{\label{distortion_a}}{\includegraphics[width=0.32\textwidth]{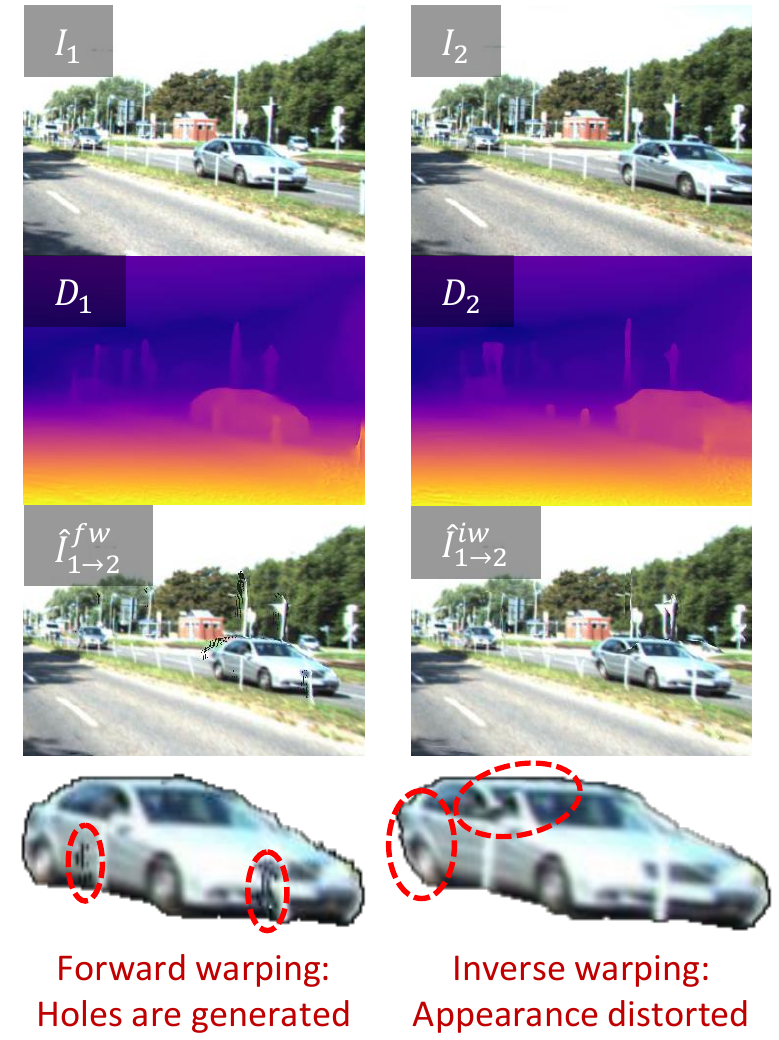}}
	\hspace{+0.5mm}
	\subcaptionbox{\label{distortion_b}}{\includegraphics[width=0.32\textwidth]{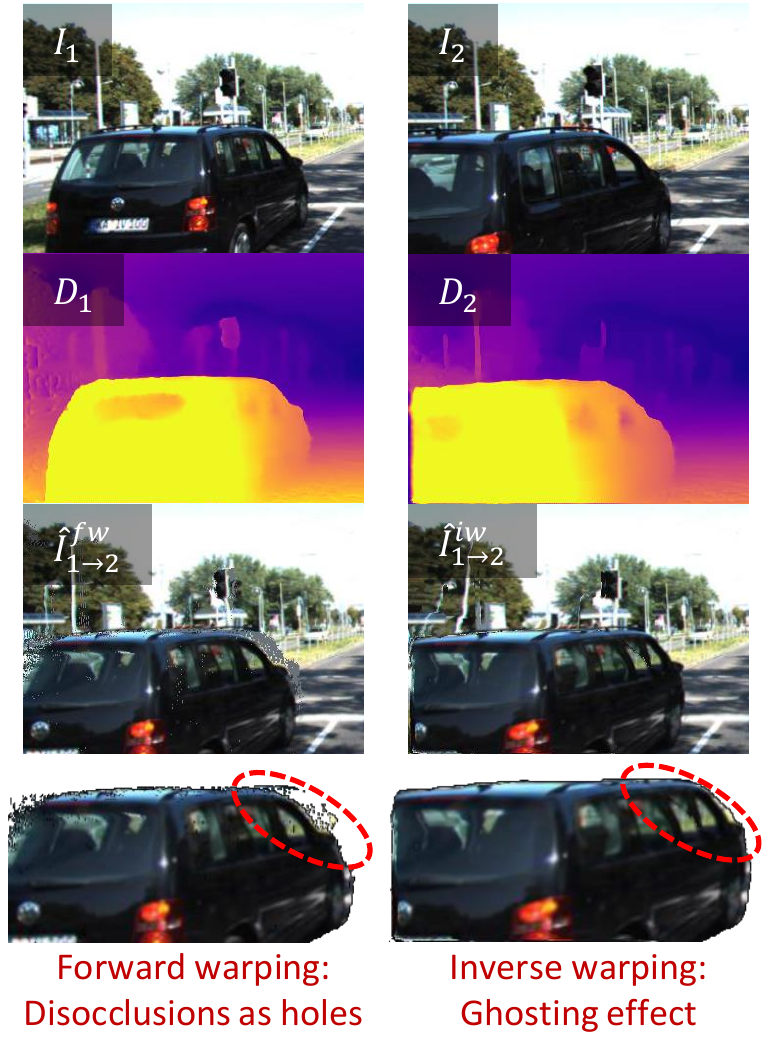}}        
\end{tabular}
\vspace{-4mm}
\caption{(a) Warping discrepancy occurs for inverse projection of moving objects. Different warping results on (b) a moving object and (c) a close object. $\hat{I}^{iw}$ and $\hat{I}^{fw}$ are warped only by the camera ego-motion.}
\vspace{-5mm}
\label{distortion}
\end{figure}

A fully differentiable warping function enables learning of structure-from-motion tasks.
This operation is first proposed by \emph{spatial transformer networks} (STN)~\cite{jaderberg2015spatial}.
Previous works for learning depth and ego-motion from unlabeled videos so far follow this \emph{grid sampling} module to synthesize adjacent views.
To synthesize $\hat{I}_{1 \rightarrow 2}$ from $I_1$, the homogeneous coordinates, $p_2$, of a pixel in $I_2$ are projected to $p_1$ as follows:
\begin{equation}
\Scale[0.99]
{
\begin{aligned}
p_{1} \sim K P^{i=0}_{2 \rightarrow 1} D_2(p_2) K^{-1}p_2 .
\end{aligned}
}
\label{eq_inv_warp}
\end{equation}
As expressed in the equation, this operation computes $\hat{I}_{1 \rightarrow 2}$ by taking the value of the homogeneous coordinates $p_1$ from the inverse rigid projection using $P^{i=0}_{2 \rightarrow 1}$ and $D_2(p_2)$.
As a result, the coordinates $p_1$ are not valid if $p_2$ lies on an object that moves between $I_1$ and $I_2$.
Therefore, the inverse warping is not suitable for removing the effects of ego-motion in dynamic scenes.
As shown in \figref{warping}, the inverse warping causes pixel discrepancy on a moving object, since it reprojects the point ($X^b_{tgt}$) from the target geometry where the 3D point has moved.
This causes distortion of the appearance of moving objects as in \figref{distortion_a} and ghosting effects~\cite{janai2018unsupervised} on the object near to the camera as in \figref{distortion_b}.
To solve this problem, we define an intermediate frame which is transformed by camera motion with reference geometry, and mitigate the residual displacement (\textcolor{orange}{orange arrow} in \figref{warping}) by training Obj-PoseNet as a supervisory signal.
In~\tabref{tab_comp}, we describe the difference between input resources of inverse and forward warping, as well as their advantages and disadvantages.

\begin{table}[!h]
\renewcommand{\tabcolsep}{1mm}
\centering
\vspace{-8mm}
\caption{Comparisons between inverse and forward warping.}
\vspace{-0mm}
\begin{adjustbox}{width=0.70\textwidth}
    \begin{tabular}{l@{\hskip 2mm}l@{\hskip 2mm}l@{\hskip 2mm}l@{\hskip 2mm}}
        \Xhline{3\arrayrulewidth}
        & Inverse warping & Forward warping \\ 
        \Xhline{1\arrayrulewidth}
        Inputs               & $I_{ref}$, $D_{tgt}$, $P^{ego}_{tgt\rightarrow ref}$  &  $I_{ref}$, $D_{ref}$, $P^{ego}_{ref\rightarrow tgt}$  \\ 
        Pros. 	             & Dense registration by grid sampling. & Geometry corresponds to reference.  \\ 
        Cons.          		 & Errors induced on moving objects.      & Holes are generated.  \\
        \Xhline{3\arrayrulewidth}
	\end{tabular}
\end{adjustbox}
\vspace{-4mm}
\label{tab_comp}
\end{table}

\begin{figure}[t]
\centering
\includegraphics[width=0.99\textwidth]{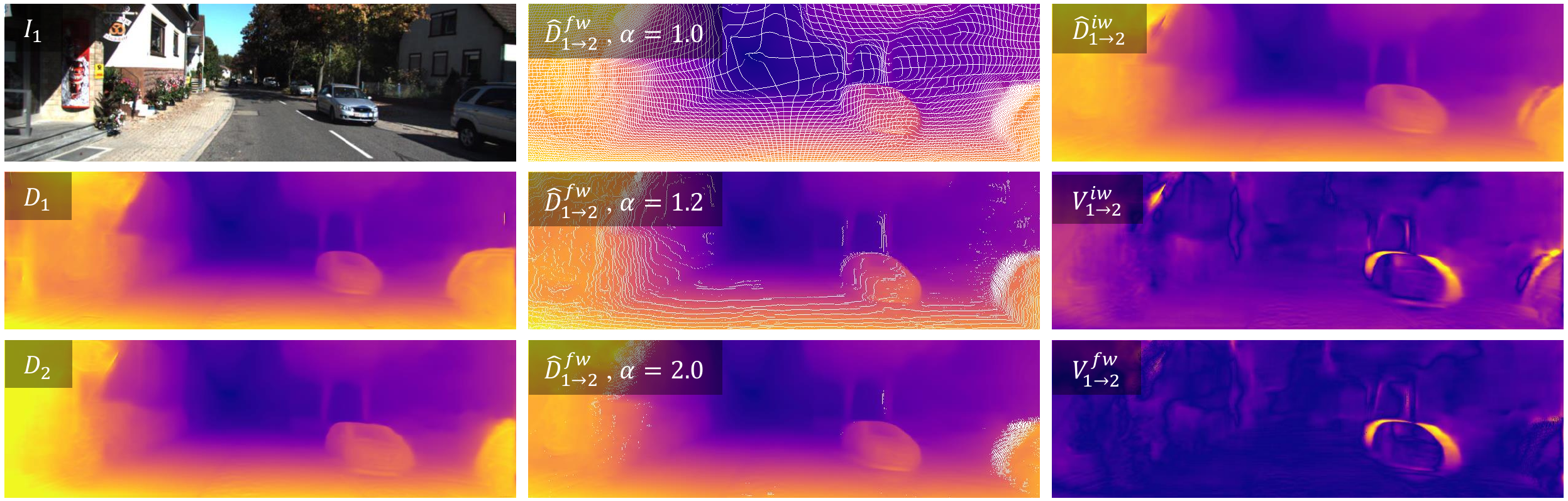}
\vspace{-2mm}
\caption{Effect of pre-upsampling reference depth for forward projection. The middle column shows the results of forward warped depth ($D^{fw}_{1 \rightarrow 2}$) with different upsampling factors, $\alpha$.
}
\vspace{-5mm}
\label{upsample}
\end{figure}

In order to synthesize the novel view (from $I_1$ to $\hat{I}_{1 \rightarrow 2}$) properly when there exist moving objects, we propose \emph{forward projective geometry}, $\mathcal{F}_{fw}(I_i, D_i, P_{i \rightarrow j}, K) \rightarrow \hat{I}_{i \rightarrow j}$ as follows:
\begin{equation}
\Scale[0.99]
{
\begin{aligned}
p_{2} \sim K P^{i=0}_{1 \rightarrow 2} D^{\uparrow}_1(p_1) (K^{\uparrow})^{-1} p_1 .
\end{aligned}
}
\label{eq_fwd_warp}
\end{equation}
Unlike inverse projection in \eqnref{eq_inv_warp}, this warping process cannot be sampled by the STN since the projection is in the forward direction (inverse of \emph{grid sampling}).
In order to make this operation differentiable, we first use sparse tensor coding to index the homogeneous coordinates $p_2$ of a pixel in $I_2$.
Invalid coordinates (exiting the view where $p_2 \notin \{(x,y)| 0 \leq x<W, 0 \leq y<H\}$) of the sparse tensor are masked out.
We then convert this sparse tensor to be dense by taking the nearest neighbor value of the source pixel.
However, this process has a limitation that there exist irregular holes due to the sparse tensor coding.
Since we need to feed those forward projected images into the neural networks in the next step, the size of the holes should be minimized.
To fill these holes as much as possible, we pre-upsample the depth map~$D^{\uparrow}_1(p_1)$ of the reference frame.
If the depth map is upsampled by a factor of $\alpha$, the camera intrinsic matrix is also upsampled as follows:
\vspace{-1mm}
\begin{equation}
\Scale[0.93]
{
\begin{aligned}
K^{\uparrow} = \begin{bmatrix}
\alpha f_x & 0          & \alpha W \\
0          & \alpha f_y & \alpha H \\
0          & 0          & 1
\end{bmatrix},
\end{aligned}
}
\label{eq_intrinsics}
\vspace{-1mm}
\end{equation}
where $(f_x, f_y)$ are focal lengths along the $x$- and $y$-axis.
\figref{upsample} shows the effect of pre-upsampling reference depth while forward warping. With an upsampling factor of $\alpha=2$ during forward projection, the holes are filled properly. 
In the following subsection, we describe the steps of how to synthesize novel views with inverse and forward projection in each instance region.


\vspace{-2mm}
\subsection{Instance-wise View Synthesis and Geometric Consistency}
\label{sec:3_3}
\vspace{-2mm}

\noindent\textbf{Instance-wise reconstruction}~
Each step of the instance-wise view synthesis is depicted in \figref{overview}.
To synthesize a novel view in an instance-wise manner, we first decompose the image region into background and object (potentially moving) regions.
With given instance masks $(M^{i}_1, M^{i}_2)$, the background mask along frames $(I_1, I_2)$ is generated as
\vspace{-2mm}
\begin{equation}
\Scale[0.94]
{
\begin{aligned}
M^{i=0}_{1,2} = (1 - \cup_{k \in \{1,2,...,n\}} M^{i=k}_1) \cap (1 - \cup_{k \in \{1,2,...,n\}} M^{i=k}_2) .
\end{aligned}
}
\label{eq_bg_mask}
\vspace{-2mm}
\end{equation}
The background mask is element-wise multiplied ($\odot$, Hadamard product) to the input frames $(I_1, I_2)$, and then concatenated along the channel axis, which is an input to the Ego-PoseNet. The camera ego-motion is computed as
\vspace{-2mm}
\begin{equation}
\Scale[0.99]
{
\begin{aligned}
P^{i=0}_{1 \rightarrow 2}, P^{i=0}_{2 \rightarrow 1} =
    \mathcal{E}_\phi( M^{i=0}_{1,2} \odot I_1, M^{i=0}_{1,2} \odot I_2).
\end{aligned}
}
\label{eq_ego_pose}
\vspace{-2mm}
\end{equation}
To learn the object motions, we first apply the forward warping, $\calF_{fw}(\cdot)$, to generate ego-motion-eliminated warped images and masks as follows:
\vspace{-1mm}
\begin{equation}
\Scale[0.99]
{
\begin{aligned}
\hat{I}^{fw}_{1 \rightarrow 2} = \calF_{fw}(I_1, D^{\uparrow}_1, P^{i=0}_{1 \rightarrow 2}, K^{\uparrow}) ,
\end{aligned}
}
\label{eq_fwd_warp}
\vspace{-2mm}
\end{equation}

\begin{equation}
\Scale[0.99]
{
\begin{aligned}
\hat{M}^{fw}_{1 \rightarrow 2} = \calF_{fw}(M_1, D^{\uparrow}_1, P^{i=0}_{1 \rightarrow 2}, K^{\uparrow}) ,
\end{aligned}
}
\label{eq_fwd_warp_mask}
\vspace{-1mm}
\end{equation}
where both equations are applied in the backward direction as well by exchanging the subscripts ${}_1$ and ${}_2$.
Now we can generate forward-projected instance images as $\hat{I}^{fw, i=k}_{1 \rightarrow 2} = \hat{M}^{fw,i=k}_{1 \rightarrow 2} \odot \hat{I}^{fw}_{1 \rightarrow 2} $ and $\hat{I}^{fw, i=k}_{2 \rightarrow 1} = \hat{M}^{fw,i=k}_{2 \rightarrow 1} \odot \hat{I}^{fw}_{2 \rightarrow 1} $.
For every object instance in the image, Obj-PoseNet predicts the $k^{th}$ object motion as
\vspace{-1mm}
\begin{equation}
\Scale[0.99]
{
\begin{aligned}
P^{i=k}_{1 \rightarrow 2}, P^{i=k}_{2 \rightarrow 1} =
    \mathcal{O}_\psi(\hat{I}^{fw, i=k}_{1 \rightarrow 2}, M^{i=k}_2 \odot I_2) ,
\end{aligned}
}
\label{eq_obj_pose}
\vspace{-1mm}
\end{equation}
where both object motions are composed of six-dimensional SE(3) translation and rotation vectors.
We merge all instance regions to synthesize the novel view.
In this step, we utilize inverse warping, $\calF_{iw}(\cdot)$.
First, the background region is reconstructed as
\vspace{-1mm}
\begin{equation}
\Scale[0.99]
{
\begin{aligned}
\hat{I}^{iw, i=0}_{1 \rightarrow 2} = M^{i=0}_{1,2} \odot \calF_{iw}(I_1, D_2, P^{i=0}_{2 \rightarrow 1}, K) ,
\end{aligned}
}
\label{eq_bg_full}
\vspace{-1mm}
\end{equation}
where the gradients are propagated with respect to $\theta$ and $\phi$.
Second, the inverse-warped $k^{th}$ instance region is represented as
\vspace{-1mm}
\begin{equation}
\Scale[0.99]
{
\begin{aligned}
\hat{I}^{fw \rightarrow iw, i=k}_{1 \rightarrow 2} =
     \calF_{iw}(\hat{I}^{fw,i=k}_{1 \rightarrow 2}, D_2, P^{i=k}_{2 \rightarrow 1}, K) ,
\end{aligned}
}
\label{eq_obj_full}
\vspace{-1mm}
\end{equation}
where the gradients are propagated with respect to $\theta$ and $\psi$.
Finally, our instance-wise fully reconstructed novel view is formulated as
\vspace{-2mm}
\begin{equation}
\Scale[0.99]
{
\begin{aligned}
\hat{I}_{1 \rightarrow 2} =
     \hat{I}^{iw, i=0}_{1 \rightarrow 2} + \sum\limits_{k \in \{1,2,...,n\}} \hat{I}^{fw \rightarrow iw, i=k}_{1 \rightarrow 2} .
\end{aligned}
}
\label{eq_full}
\vspace{-2mm}
\end{equation}
Note that the above three equations are applied in either the forward or backward directions by switching the subscripts ${}_1$ and ${}_2$.

\noindent\textbf{Instance-wise mini-batch re-arrangement}~
While training Obj-PoseNet, the number of instance images may change after each iteration.
In order to avoid inefficient iterative training, we fix the maximum number of instances per image (sampled in order of instance size) and re-arrange the mini-batches with respect to the total number of instances in the mini-batch.
For example, if the mini-batch has four frames and each frame contains $\{3, 1, 0, 2\}$ instances, then the rearranged mini-batch size is $3+1+0+2=6$.
The gradients for individual instances are normalized by the number of pixels in an object, as well as the number of instances (rearranged batch size).

\noindent\textbf{Instance mask propagation}~
Through the process of forward and inverse warping, the instance mask is also propagated to contain the information of instance position and pixel validity.
In the case of the $k^{th}$ instance mask $M^{i=k}_1$, the forward and inverse warped mask is expressed as follows:
\vspace{-1mm}
\begin{equation}
\Scale[0.99]
{
\begin{aligned}
\hat{M}^{fw \rightarrow iw, i=k}_{1 \rightarrow 2} =
     \calF_{iw}(\hat{M}^{fw,i=k}_{1 \rightarrow 2}, D_2, P^{i=k}_{2 \rightarrow 1}, K) .
\end{aligned}
}
\label{eq_inst_mask}
\vspace{-1mm}
\end{equation}
Note that the forward warped mask $\hat{M}^{fw,i=k}_{1 \rightarrow 2}$ has holes due to the sparse tensor coding.
To keep the binary format and avoid interpolation near the holes while inverse warping, we round up the fractional values after each warping operation.
The final valid instance mask is expressed as follows:
\vspace{-2mm}
\begin{equation}
\Scale[0.99]
{
\begin{aligned}
\hat{M}_{1 \rightarrow 2} =
     M^{i=0}_{1,2} + \sum\limits_{k \in \{1,2,...,n\}} \hat{M}^{fw \rightarrow iw, i=k}_{1 \rightarrow 2} .
\end{aligned}
}
\label{eq_mask_sum}
\vspace{-2mm}
\end{equation}


\noindent\textbf{Instance-wise geometric consistency}~
We impose the geometric consistency loss for each region of an instance.
Following the work by Bian~\etal~\cite{bian2019unsupervised}, we constrain the geometric consistency during inverse warping.
With the predicted depth map and warped instance mask, $D_1$ can be spatially aligned to the frame $D_2$ by forward and inverse warping, represented as $M^{i=0}_{1,2} \odot \hat{D}^{iw, i=0}_{1 \rightarrow 2}$ and $\hat{M}^{fw \rightarrow iw, i=k}_{1 \rightarrow 2} \odot \hat{D}^{fw \rightarrow iw, i=k}_{1 \rightarrow 2}$ respectively for background and instance regions.
In addition, $D_2$ can be scale-consistently transformed to the frame $D_1$, represented as $M^{i=0}_{1,2} \odot D^{sc, i=0}_{2 \rightarrow 1}$ and $\hat{M}^{fw \rightarrow iw, i=k}_{1 \rightarrow 2} \odot D^{sc, i=k}_{2 \rightarrow 1}$ respectively for background and instance regions.
Based on this instance-wise operation, we compute the unified depth inconsistency map as:
\vspace{-2mm}
\begin{equation}
\Scale[0.92]
{
\begin{aligned}
D^{diff, i=0}_{1 \rightarrow 2} = M^{i=0}_{1,2} \odot
     \frac{ |\hat{D}^{iw, i=0}_{1 \rightarrow 2} - D^{sc, i=0}_{2 \rightarrow 1}| }
          {  \hat{D}^{iw, i=0}_{1 \rightarrow 2} + D^{sc, i=0}_{2 \rightarrow 1}  } ,
\end{aligned}
}
\label{eq_gc_bg}
\vspace{-2mm}
\end{equation}

\begin{equation}
\Scale[0.92]
{
\begin{aligned}
D^{diff, i=k}_{1 \rightarrow 2} = \hat{M}^{fw \rightarrow iw, i=k}_{1 \rightarrow 2} \odot
     \frac{ |\hat{D}^{fw \rightarrow iw, i=k}_{1 \rightarrow 2} - D^{sc, i=k}_{2 \rightarrow 1}| }
          {  \hat{D}^{fw \rightarrow iw, i=k}_{1 \rightarrow 2} + D^{sc, i=k}_{2 \rightarrow 1}  } ,
\end{aligned}
}
\label{eq_gc_obj}
\vspace{-2mm}
\end{equation}
where each line is applied to either the background or an instance region, and both are applied in either the forward and backward directions by exchanging the subscripts ${}_1$ and ${}_2$.
Note that the above depth inconsistency maps are spatially aligned to the frame $D_2$.
Therefore, we can integrate the depth inconsistency maps from the background and instance regions as follows:
\vspace{-2mm}
\begin{equation}
\Scale[0.99]
{
\begin{aligned}
D^{diff}_{1 \rightarrow 2} = D^{diff, i=0}_{1 \rightarrow 2} + \sum\limits_{k \in \{1,2,...,n\}} D^{diff, i=k}_{1 \rightarrow 2} .
\end{aligned}
}
\label{eq_gc}
\vspace{-2mm}
\end{equation}

\noindent\textbf{Training loss}~
In order to handle occluded, view-exiting, and invalid instance regions, we leverage \eqnref{eq_mask_sum} and \eqnref{eq_gc}.
We generate a weight mask as $1-D^{diff}_{1 \rightarrow 2}$ and this is multiplied to the valid instance mask $\hat{M}_{1 \rightarrow 2}$.
Finally, our weighted valid mask is formulated as:
\vspace{-1mm}
\begin{equation}
\Scale[0.99]
{
\begin{aligned}
V_{1 \rightarrow 2} = (1-D^{diff}_{1 \rightarrow 2}) \odot \hat{M}_{1 \rightarrow 2} .
\end{aligned}
}
\label{eq_valid}
\vspace{-0mm}
\end{equation}
The reconstruction loss $\calL_r$ is expressed as follows:
\vspace{-1mm}
\begin{equation}
\Scale[0.86]
{
\begin{aligned}
\calL_r =
\sum\limits_{{\rm{x}} \in X} V_{1 \rightarrow 2}({\rm{x}}) \cdot  \left\{ (1-\gamma) \cdot \left\| { {I_2({\rm{x}}) - {\hat{I}_{1 \rightarrow 2}}({\rm{x}})} } \right\|_{1} + \gamma \left( 1 - SSIM(I_2({\rm{x}}), {{\hat I}_{1 \rightarrow 2}}({\rm{x}})) \right)  \right\} ,
\end{aligned}
}
\label{eq_loss_recon}
\vspace{-1mm}
\end{equation}
where $\rm{x}$ is the location of each pixel, $SSIM(\cdot)$ is the structural similarity index~\cite{wang2004image}, and $\gamma$ is set to 0.8 based on cross-validation.
The geometric consistency loss $\calL_g$ is expressed as follows:
\vspace{-1mm}
\begin{equation}
\Scale[0.99]
{
\begin{aligned}
\calL_g = \sum\limits_{{\rm{x}} \in X} \hat{M}_{1 \rightarrow 2}({\rm{x}}) \cdot D^{diff}_{1 \rightarrow 2}({\rm{x}}).
\end{aligned}
}
\label{eq_loss_gc}
\vspace{-1mm}
\end{equation}

To mitigate spatial fluctuation, we incorporate a smoothness term to regularize the predicted depth map.
We apply the edge-aware smoothness loss proposed by Ranjan~\etal~\cite{ranjan2019collaboration}, which is described as:
\vspace{-1mm}
\begin{equation}
\Scale[0.99]
{
\begin{aligned}
\calL_s = \sum\limits_{{\rm{x}} \in X} ( \nabla D_1({\rm{x}}) \cdot e^{-\nabla I_1({\rm{x}})} )^2 .
\end{aligned}
}
\label{eq_loss_s}
\vspace{-1mm}
\end{equation}
Note that the above loss functions are imposed for both forward and backward directions by switching the subscripts ${}_1$ and ${}_2$.

Since the dataset has a low proportion of moving objects, the learned motions of objects tend to converge to zero.
The same issue has been raised in the previous study~\cite{casser2019depth}.
To supervise the approximate amount of an object's movement, we constrain the motion of the object with a translation prior.
We compute this translation prior, $t_p$, by subtracting the mean estimate of the object's 3D points in the forward warped frame into that of the target frame's 3D object points. This represents the mean estimated 3D vector of the object's motion.
The object translation constraint loss is defined as follows:
\vspace{-1mm}
\begin{equation}
\Scale[0.99]
{
\begin{aligned}
\calL_t = \sum\limits_{k \in \{1,2,...,n\}} ||t^{i=k} - t^{i=k}_p||_1 ,
\end{aligned}
}
\label{eq_loss_t}
\vspace{-3mm}
\end{equation}
where $t^{i=k}$ and $t^{i=k}_p$ are predicted object translation from Obj-PoseNet and the translation prior on the $k^{th}$ instance mask.

Although we have accounted for instance-wise geometric consistency, there still exists a trivial case of infinite depth for a moving object, which has the same motion as the camera motion, especially for vehicles in front.
To mitigate this issue, we adopt the object height constraint loss proposed by a previous study~\cite{casser2019depth}, which is described as:
\vspace{-2mm}
\begin{equation}
\Scale[0.98]
{
\begin{aligned}
\calL_h = \sum\limits_{k \in \{1,2,...,n\}} \frac{1}{\overline{D}} \cdot ||D \odot M^{i=k} - \frac{ f_y \cdot p^{i=k}_h }{ h^{i=k} }||_1 ,
\end{aligned}
}
\label{eq_loss_h}
\vspace{-3mm}
\end{equation}
where $\overline{D}$ is the mean estimated depth, and ($p^{i=k}_h$, $h^{i=k}$) are a learnable height prior and pixel height of the $k^{th}$ instance.
The final loss is a weighted summation of the five loss terms, defined as \eqnref{eq_loss}.





\vspace{-2mm}
\section{Experiments}
\label{sec:4}
\vspace{-2mm}

We evaluate the performance of our frameworks and compare with previous unsupervised methods on single view depth and visual odometry tasks.
We train and test our method on KITTI~\cite{geiger2012we} for benchmarking.

\vspace{-2mm}
\subsection{Implementation Details}
\label{sec:4_1}
\vspace{-2mm}

\noindent\textbf{Network details}~
For DepthNet, we use DispResNet~\cite{ranjan2019collaboration} based on an encoder-decoder structure. The network can generate multi-scale outputs (six different scales), but the single-scale training converges faster and produces better performance as shown from SC-SfM~\cite{bian2019unsupervised}.
The structures of Ego-PoseNet and Obj-PoseNet are the same, but the weights are not shared. They consist of seven convolutional layers and regress the relative pose as three Euler angles and three translation vectors. 

\noindent\textbf{Training}~
Our system is implemented in PyTorch~\cite{paszke2017automatic}.
We train our networks using the ADAM optimizer~\cite{kingma2015adam} with $\beta_1 = 0.9$ and $\beta_2 = 0.999$ on $4\times$Nvidia RTX 2080 GPUs.
The training image resolution is set to $832\times 256$ and the video data is augmented with random scaling, cropping, and horizontal flipping.
We set the mini-batch size to 4 and train the networks over 200 epochs with 1,000 randomly sampled batches in each epoch.
The initial learning rate is set to $10^{-4}$ and is decreased by half every 50 epochs.
The loss weights are set to $\lambda_r = 1.0$, $\lambda_g = 0.5$, $\lambda_s = 0.05$, $\lambda_t = 0.1$, and $\lambda_h = 0.001$.

\begin{figure}[t]
\centering
\includegraphics[width=0.47\textwidth]{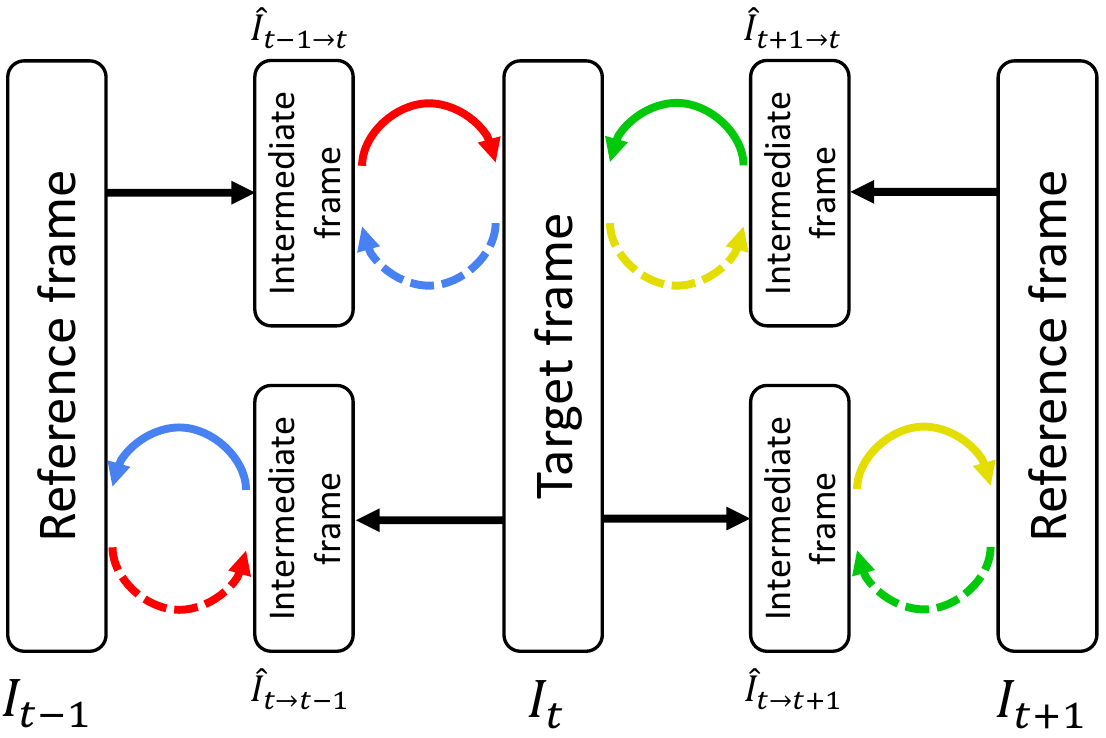}
\vspace{-3mm}
\caption{Cyclic intermediate representation while training. Intermediate frames are generated to eliminate the effect of camera motion. 
}
\label{cycle}
\vspace{-3mm}
\end{figure}

\noindent\textbf{Intermediate representation}~
While training, we take three consecutive frames as input to train our joint networks.
Our three-frame cyclic training is depicted in~\figref{cycle}.
Dynamic scenes are hard to handle with rigid projective geometry in a one-shot manner.
We utilize an intermediate frame which enables decomposition of ego-motion-driven global view synthesis and residual view synthesis by object motions.
From this, several warping directions can be proposed.
The arrows in \figref{cycle} represent the warping direction of RGB images and depth maps.
We tried to optimize Obj-PoseNet by warping to the intermediate frames (dashed arrows); however, the network did not converge.
One important point here is that we need to feed the supervisory signals generated at the original timestamps $\{I_{t-1}, I_t, I_{t+1}\}$, not in the intermediate frames $\hat{I}$.
Although we generate photometric and geometric supervision only in the reference or target frames, we utilize the object motions while warping to the intermediate frames. 
We regularize the object motions by averaging two motions in the same direction, identified with the same colored arrows.

\vspace{-2mm}
\subsection{Auto-annotation of Video Instance Segmentation Dataset}
\label{sec:4_2}
\vspace{-2mm}

We introduce an auto-annotation scheme to generate a video instance segmentation dataset from the existing driving dataset, KITTI~\cite{geiger2012we} and Cityscapes~\cite{cordts2016cityscapes}.
To this end, we adopt an off-the-shelf instance segmentation model, \eg, Mask R-CNN~\cite{he2017mask} and PANet~\cite{liu2018path}, and an optical flow model, PWC-Net~\cite{sun2018pwc}.
We first compute the instance segmentation for every image frame, and calculate the Intersection over Union (IoU) score table among instances in each frame.
The occluded and disoccluded regions are handled by bidirectional flow consensus proposed in UnFlow~\cite{meister2017unflow}.
If the maximal IoU in the adjacent frame is above a threshold ($\tau = 0.5$), then the instance is assumed as tracked and both masks are assigned with the same ID.

\begin{table}[t]
\centering
\caption{Validation of our auto-annotation for video instance segmentation on KITTI MOTS validation set (car class). 
}
\vspace{+1mm}
\begin{adjustbox}{width=0.99\textwidth}
\setlength{\tabcolsep}{5pt}
\begin{tabular}{@{}c|ccccccccc@{}}
\Xhline{4\arrayrulewidth}
Method & Backbone & pretrained & sMOTSA & MOTSA & MOTSP & IDS & TP & FP & FN \\
\Xhline{2\arrayrulewidth}
TrackR-CNN~\cite{voigtlaender2019mots}            & TrackR-CNN/ResNet-101 & COCO + MOTS & 76.2 & 87.8 & \textbf{87.2} & 93  & 7276 & \textbf{134} & 753  \\
Mask R-CNN + maskprop~\cite{voigtlaender2019mots} & Mask R-CNN/ResNet-101 & COCO + MOTS & 75.1 & 86.6 & 87.1 & --  & --   & --  & --   \\
CAMOT~\cite{ovsep2018track}                       & TrackR-CNN/ResNet-101 & COCO + MOTS & 67.4 & 78.6 & 86.5 & 220 & 6702 & 172 & 1327 \\
CIWT~\cite{osep2017combined}                      & TrackR-CNN/ResNet-101 & COCO + MOTS & 68.1 & 79.4 & 86.7 & 106 & 6815 & 333 & 1214 \\
\Xhline{2\arrayrulewidth}
\multirow{3}{*}{Ours}          & PANet/ResNet-50       & COCO        & 66.9 & 79.1 & 85.2 & (0) & 6647 & 291 & 1382 \\
                               & Mask R-CNN/ResNet-101 & COCO        & 66.1 & 78.1 & 85.5 & (0) & 6614 & 347 & 1415 \\
                               & Mask R-CNN/ResNet-101 & COCO + MOTS & 77.4 & 89.6 & 86.6 & (0) & \textbf{7338} & 142 & \textbf{691}  \\
\Xhline{4\arrayrulewidth}
\end{tabular}
\end{adjustbox}
\label{tab_mots}
\vspace{-4mm}
\end{table}

To validate the performance of our annotation, we measure extended multi-object tracking metrics on KITTI multi-object tracking and segmentation (MOTS) dataset~\cite{voigtlaender2019mots}.
Since our training scheme only needs tracking information between adjacent frames ($I_1, I_2$), the number of ID switching (IDS) is not important in our case.
Therefore, we set IDS as zero when measuring sMOTSA and MOTSA for our method.
Furthermore, considering our system, false negatives are more important than false positives since the static objects also can be interpreted by our forward projective geometry.
We show that ours achieves favorable results on the KITTI MOTS compared to the state-of-the-art work~\cite{voigtlaender2019mots} by adopting existing pretrained models (MOTS-pretrained Mask R-CNN and KITTI-pretrained PWC-Net).
Our auto-annotation tool will become publicly available.

\vspace{-2mm}
\subsection{Ablation Study}
\vspace{-2mm}

We conduct an ablation study to validate the effect of our forward projective geometry and instance-wise geometric consistency term on monocular depth estimation.
The ablation is performed with the Eigen split~\cite{eigen2014depth} of the KITTI dataset.
The models are validated with the AbsRel metric by separating the background and object areas, which are masked by our annotation.
As described in \tabref{tab_ablation}, we first evaluate SC-SfM~\cite{bian2019unsupervised} as a baseline, which is not trained with instance knowledge (the $1^{st}$ and $2^{nd}$ models).
Since there are no instance masks, DepthNet is trained by inverse warping the whole image.
With the given instance masks, we try both inverse and forward warping on the object areas. 
The inverse warping on the objects slightly improves the depth estimation; however, we observe that Obj-PoseNet does not converge (the $3^{rd}$ and $4^{th}$ models). 
Rather, the performance is degraded when using the instance-wise geometric consistency term with inverse warping on the objects.
We conjecture that the uncertainty in learning the depth of the object area degrades the performance on the background depth around which the object is moving.
However, the forward warping on the objects improves the depth estimation on both background and object areas (the $5^{th}$ and $6^{th}$ models).
This shows that well-optimized Obj-PoseNet helps to boost the performance of DepthNet and they complement each other.
We note that the background is still inverse warped to synthesize the target view and the significant performance improvement comes from the instance-wise geometric loss incorporated with forward projection while warping the object areas.


\begin{table}[t]
\centering
\caption{Ablation study on KITTI Eigen split of the monocular depth estimation for both background (bg.) and object (obj.) areas. 
}
\vspace{+1mm}
\begin{adjustbox}{width=0.59\textwidth}
\setlength{\tabcolsep}{5pt}
\begin{tabular}{ccccccc}
\Xhline{3\arrayrulewidth}
\multirow{2}[3]{*}{\shortstack{Instance \\ knowledge}} & \multirow{2}[3]{*}{\shortstack{Geometric \\ consistency}} & \multicolumn{2}{c}{Object warping} & \multicolumn{3}{c}{AbsRel} \\
\cmidrule(l{2pt}r{2pt}){3-4} \cmidrule(l{2pt}r{2pt}){5-7}
&  & inverse & forward & all & bg. & obj.  \\
\Xhline{2\arrayrulewidth}
\xmark & \xmark & \xmark & \xmark & 0.156 & 0.142 & 0.396 \\
\xmark & \cmark & \xmark & \xmark & \underbar{0.137} & \underbar{0.124} & 0.309 \\
\cmark & \xmark & \cmark & \xmark & 0.151 & 0.138 & 0.377 \\
\cmark & \cmark & \cmark & \xmark & 0.146 & 0.131 & 0.362 \\
\cmark & \xmark & \xmark & \cmark & 0.143 & 0.133 & \underbar{0.285} \\
\rowcolor{red!7}
\cmark & \cmark & \xmark & \cmark & \textbf{0.124} & \textbf{0.119} & \textbf{0.178} \\
\Xhline{3\arrayrulewidth}
\end{tabular}
\end{adjustbox}
\label{tab_ablation}
\vspace{-5mm}
\end{table}

\begin{table}[t]
    \centering
    \caption{Monocular depth estimation results on the KITTI Eigen test split. Models trained on KITTI raw dataset are denoted by `K'. Models pretrained on Cityscapes and fine-tuned on KITTI are denoted by `CS+K'. `M' and `S' denote monocular and stereo setup for training, `D' denotes depth supervision. \textbf{Bold}: Best, \underbar{Underbar}: Second best.}
    \vspace{+1mm}
    \begin{adjustbox}{width=0.93\textwidth}
    \setlength{\tabcolsep}{5pt}
    \begin{tabular}{lccccccccc}

    \Xhline{5\arrayrulewidth}
    \multirow{2}{*}{Method} & \multirow{2}{*}{Training} & \multicolumn{4}{c}{Error metric $\downarrow$} & & \multicolumn{3}{c}{Accuracy metric $\uparrow$}  \\
    \cline{3-6} \cline{8-10}
     &  & AbsRel  & SqRel & RMSE & RMSE log & & $\delta < 1.25$ & $\delta < 1.25^2$ & $\delta < 1.25^3$ \\ \Xhline{3\arrayrulewidth}

    Eigen~\etal~\cite{eigen2014depth}           & K (M+D) & 0.203 & 1.548 & 6.307 & 0.282 & & 0.702 & 0.890 & 0.958 \\
    Liu~\etal~\cite{liu2016learning}            & K (S+D) & 0.202 & 1.614 & 6.523 & 0.275 & & 0.678 & 0.895 & 0.965 \\
    Klodt~\etal~\cite{klodt2018supervising}     & K (M+D) & 0.166 & 1.490 & 5.998 & --    & & 0.778 & 0.919 & 0.966 \\
    Kuznietsov~\etal~\cite{kuznietsov2017semi}  & K (S+D) & 0.113 & 0.741 & 4.621 & 0.189 & & 0.862 & 0.960 & 0.986 \\
    DORN~\cite{fu2018deep}                      & K (M+D) & 0.072 & 0.307 & 2.727 & 0.120 & & 0.932 & 0.984 & 0.994 \\
    \Xhline{2\arrayrulewidth}
    
    Garg~\etal~\cite{garg2016unsupervised}      & K (S) & 0.152 & 1.226 & 5.849 & 0.246 &  & 0.784 & 0.921 & 0.967 \\
    Godard~\etal~\cite{godard2017unsupervised}  & K (S) & 0.148 & 1.344 & 5.927 & 0.247 &  & 0.803 & 0.922 & 0.964 \\
    Zhan~\etal~\cite{zhan2018unsupervised}      & K (S) & 0.144 & 1.391 & 5.869 & 0.241 &  & 0.803 & 0.928 & 0.969 \\
    Monodepth2~\cite{godard2019digging}         & K (S) & 0.130 & 1.144 & 5.485 & 0.232 &  & 0.831 & 0.932 & 0.968 \\
    \Xhline{2\arrayrulewidth}

    SfM-Learner~\cite{zhou2017unsupervised}     & K (M) & 0.208 & 1.768 & 6.856 & 0.283 &  & 0.678 & 0.885 & 0.957 \\
    Yang~\etal~\cite{yang2017unsupervised}      & K (M) & 0.182 & 1.481 & 6.501 & 0.267 &  & 0.725 & 0.906 & 0.963 \\
    Wang~\etal~\cite{wang2018learning}          & K (M) & 0.151 & 1.257 & 5.583 & 0.228 &  & 0.810 & 0.936 & 0.974 \\
    Mahjourian~\etal~\cite{mahjourian2018unsupervised} & K (M) & 0.163 & 1.240 & 6.220 & 0.250 &  & 0.762 & 0.916 & 0.968 \\
    LEGO~\etal~\cite{yang2018lego}              & K (M) & 0.162 & 1.352 & 6.276 & 0.252 &  & 0.783 & 0.921 & 0.969 \\
    GeoNet~\cite{yin2018geonet}                 & K (M) & 0.155 & 1.296 & 5.857 & 0.233 &  & 0.793 & 0.931 & 0.973 \\
    DF-Net~\cite{zou2018df}                     & K (M) & 0.150 & 1.124 & 5.507 & 0.223 &  & 0.806 & 0.933 & 0.973 \\
    DDVO~\cite{wang2018learning}                & K (M) & 0.151 & 1.257 & 5.583 & 0.228 &  & 0.810 & 0.936 & 0.974 \\
    CC~\cite{ranjan2019collaboration}           & K (M) & 0.140 & 1.070 & 5.326 & 0.217 &  & 0.826 & 0.941 & 0.975 \\
    SC-SfM~\cite{bian2019unsupervised}          & K (M) & 0.137 & 1.089 & 5.439 & 0.217 &  & 0.830 & 0.942 & 0.975 \\
    Struct2Depth~\cite{casser2019depth}         & K (M) & 0.141 & 1.026 & 5.290 & 0.215 &  & 0.816 & \underbar{0.945} & \underbar{0.979} \\
    GLNet~\cite{chen2019self}                   & K (M) & 0.135 & 1.070 & 5.230 & \underbar{0.210} &  & \underbar{0.841} & \textbf{0.948} & \textbf{0.980} \\
    Monodepth2~\cite{godard2019digging}         & K (M) & \underbar{0.132} & \underbar{1.044} & \textbf{5.142} & \underbar{0.210} &  & \textbf{0.845} & \textbf{0.948} & 0.977 \\
    \rowcolor{red!7}
    Ours                                        & K (M) & \textbf{0.124} & \textbf{1.009} & \underbar{5.176} & \textbf{0.208} &  & 0.839 & 0.942 & \textbf{0.980}  \\
    \Xhline{2\arrayrulewidth}
    
    SfM-Learner~\cite{zhou2017unsupervised}     & CS+K (M) & 0.198 & 1.836 & 6.565 & 0.275 &  & 0.718 & 0.901 & 0.960 \\
    Yang~\etal~\cite{yang2017unsupervised}      & CS+K (M) & 0.165 & 1.360 & 6.641 & 0.248 &  & 0.750 & 0.914 & 0.969 \\
    Wang~\etal~\cite{wang2018learning}          & CS+K (M) & 0.151 & 1.257 & 5.583 & 0.228 &  & 0.810 & 0.936 & 0.974 \\
    Mahjourian~\etal~\cite{mahjourian2018unsupervised} & CS+K (M) & 0.159 & 1.231 & 5.912 & 0.243 &  & 0.784 & 0.923 & 0.970 \\
    GeoNet~\cite{yin2018geonet}                 & CS+K (M) & 0.153 & 1.328 & 5.737 & 0.232 &  & 0.802 & 0.934 & 0.972 \\
    DF-Net~\cite{zou2018df}                     & CS+K (M) & 0.146 & 1.182 & 5.215 & 0.213 &  & 0.818 & 0.943 & 0.978 \\
    CC~\cite{ranjan2019collaboration}           & CS+K (M) & 0.139 & \underbar{1.032} & \underbar{5.199} & 0.213 &  & 0.827 & 0.943 & 0.977 \\
    SC-SfM~\cite{bian2019unsupervised}          & CS+K (M) & \underbar{0.128} & 1.047 & 5.234 & \underbar{0.208} &  & \underbar{0.846} & \underbar{0.947} & \underbar{0.976}\\
    \rowcolor{red!7}
    Ours                                        & CS+K (M) & \textbf{0.119} & \textbf{0.985} & \textbf{5.021} & \textbf{0.204} &  & \textbf{0.856} & \textbf{0.949} & \textbf{0.981}  \\

    \Xhline{5\arrayrulewidth}

    \end{tabular}
    \end{adjustbox}
    \label{tab_kitti}
    \vspace{-4mm}
\end{table}

\vspace{-2mm}
\subsection{Monocular Depth Estimation}
\vspace{-2mm}

\noindent\textbf{Test setup}~
Following the test setup proposed for SfM-Learner~\cite{zhou2017unsupervised}, we train and test our models with the Eigen split~\cite{eigen2014depth} of the KITTI dataset.
We compare the performance of the proposed method with recent state-of-the-art works~\cite{chen2019self,casser2019depth,bian2019unsupervised,ranjan2019collaboration,godard2019digging} for unsupervised single-view depth estimation.

\begin{figure}[t]
\centering
\includegraphics[width=0.99\textwidth]{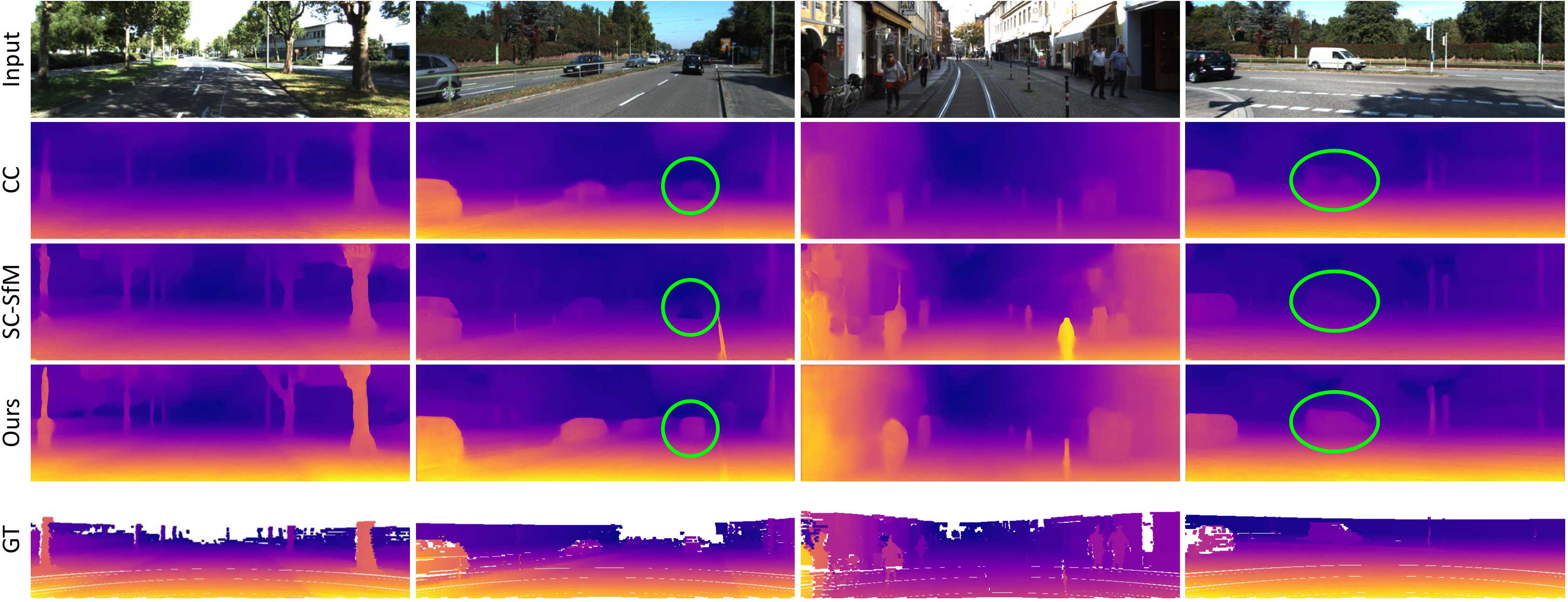}
\vspace{-2mm}
\caption{Qualitative results of single-view depth prediction on the KITTI test set (from the Eigen split). Green circles highlight favorable results of our method for depth prediction on moving objects.}
\label{depth}
\vspace{-5mm}
\end{figure}

\noindent\textbf{Results analysis}~
\tabref{tab_kitti} shows the results on the KITTI Eigen split test, where the proposed method achieves state-of-the-art performance in the single view depth prediction task with unsupervised training. 
The advantage is evident from using instance masks and constraining the instance-wise photometric and geometric consistencies.
Note that we do not need instance masks in the test phase for DepthNet.

We show qualitative results on single-view depth estimation in \figref{depth}.
The compared methods are CC~\cite{ranjan2019collaboration} and SC-SfM~\cite{bian2019unsupervised}, which have the same network structure (DispResNet) for depth map prediction.
Ours produces better depth representations on moving objects than the previous methods.
As the previous studies do not consider dynamics of objects when finding pixel correspondences, their results of training on object distance could be either farther or closer than the actual distance.
This is a traditional limitation for the task of self-supervised learning of depth from monocular videos; however, our networks self-disentangle moving and static object regions by our instance-wise losses.

\vspace{-2mm}
\subsection{Visual Odometry Estimation}
\vspace{-2mm}

\noindent\textbf{Test setup}~
We evaluate the performance of our Ego-PoseNet on the KITTI visual odometry dataset.
Following the evaluation setup of SfM-Learner~\cite{zhou2017unsupervised}, we use the sequences 00-08 for training, and sequences 09 and 10 for tests.
In our case, since the potentially moving object masks are fed together with the image sequences while training Ego-PoseNet, we test the performance of visual odometry under two conditions: with and without instance masks.

\noindent\textbf{Results analysis}~
\tabref{tab_odom} shows the results on the KITTI visual odometry test.
We measure the Absolute Trajectory Error (ATE) and achieve state-of-the-art performance.
Although we do not use the instance mask, the result of sequence 10 produces favorable performance.
This is because the scene does not have many potentially moving objects, \eg, vehicles and pedestrians, so the result is not affected much by using or not using instance masks.

We show the qualitative result on the sequence 09 in \figref{vo}. 
For the fair comparison, the PoseNet models from GeoNet~\cite{yin2018geonet} and SC-SfM~\cite{bian2019unsupervised} have the same architecture with the Ego-PoseNet, and all three models are trained on KITTI raw dataset.
Although we do not use the instance mask while testing, ours visually achieves the best result.

\begin{table}[t]
\begin{minipage}{.62\linewidth}
    \centering
    \renewcommand{\tabcolsep}{1mm}
    \captionof{table}{Absolute Trajectory Error (ATE) on the KITTI visual odometry dataset (lower is better). \textbf{Bold}: Best.}
    \vspace{+3mm}
    \begin{adjustbox}{width=0.99\textwidth}
    \begin{tabular}{lccc}
    \Xhline{4\arrayrulewidth}
    Method & No. frames & Seq. 09 & Seq. 10 \\
    \Xhline{1\arrayrulewidth}
    ORB-SLAM (full)~\cite{mur2015orb}           & All & $0.014\pm0.008$ & $0.012\pm0.011$ \\
    ORB-SLAM (short)~\cite{mur2015orb}          & 5   & $0.064\pm0.141$ & $0.064\pm0.130$ \\
    SfM-Learner~\cite{zhou2017unsupervised}     & 5   & $0.021\pm0.017$ & $0.020\pm0.015$ \\
    SfM-Learner (updated)~\cite{zhou2017unsupervised}    & 5   & $0.016\pm0.017$ & $0.013\pm0.015$ \\
    DF-Net~\cite{zou2018df}                     & 3   & $0.017\pm0.007$ & $0.015\pm0.009$ \\
    Vid2Depth~\cite{mahjourian2018unsupervised} & 3   & $0.013\pm0.017$ & $0.012\pm0.015$ \\
    GeoNet~\cite{yin2018geonet}                 & 3   & $0.012\pm0.007$ & $0.012\pm0.009$ \\
    CC~\cite{ranjan2019collaboration}           & 3   & $0.012\pm0.007$ & $0.012\pm0.008$ \\
    Struct2Depth~\cite{casser2019depth}         & 3   & $0.011\pm0.006$ & $0.011\pm0.010$ \\
    GLNet~\cite{chen2019self}                   & 3   & $0.011\pm0.006$ & $0.011\pm0.009$ \\
    \Xhline{1\arrayrulewidth}
    Ours (w/o inst.)                            & 3   & $0.012\pm0.008$ & $0.011\pm0.010$ \\
    Ours (w/ inst.)                             & 3   & $\mathbf{0.010\pm0.013}$ & $\mathbf{0.011\pm0.008}$ \\
    \Xhline{4\arrayrulewidth}
    \end{tabular}
    \end{adjustbox}
    \label{tab_odom}
\end{minipage}
\hspace{+0mm}
\begin{minipage}{.37\linewidth}
    \centering
	\includegraphics[width=0.99\textwidth]{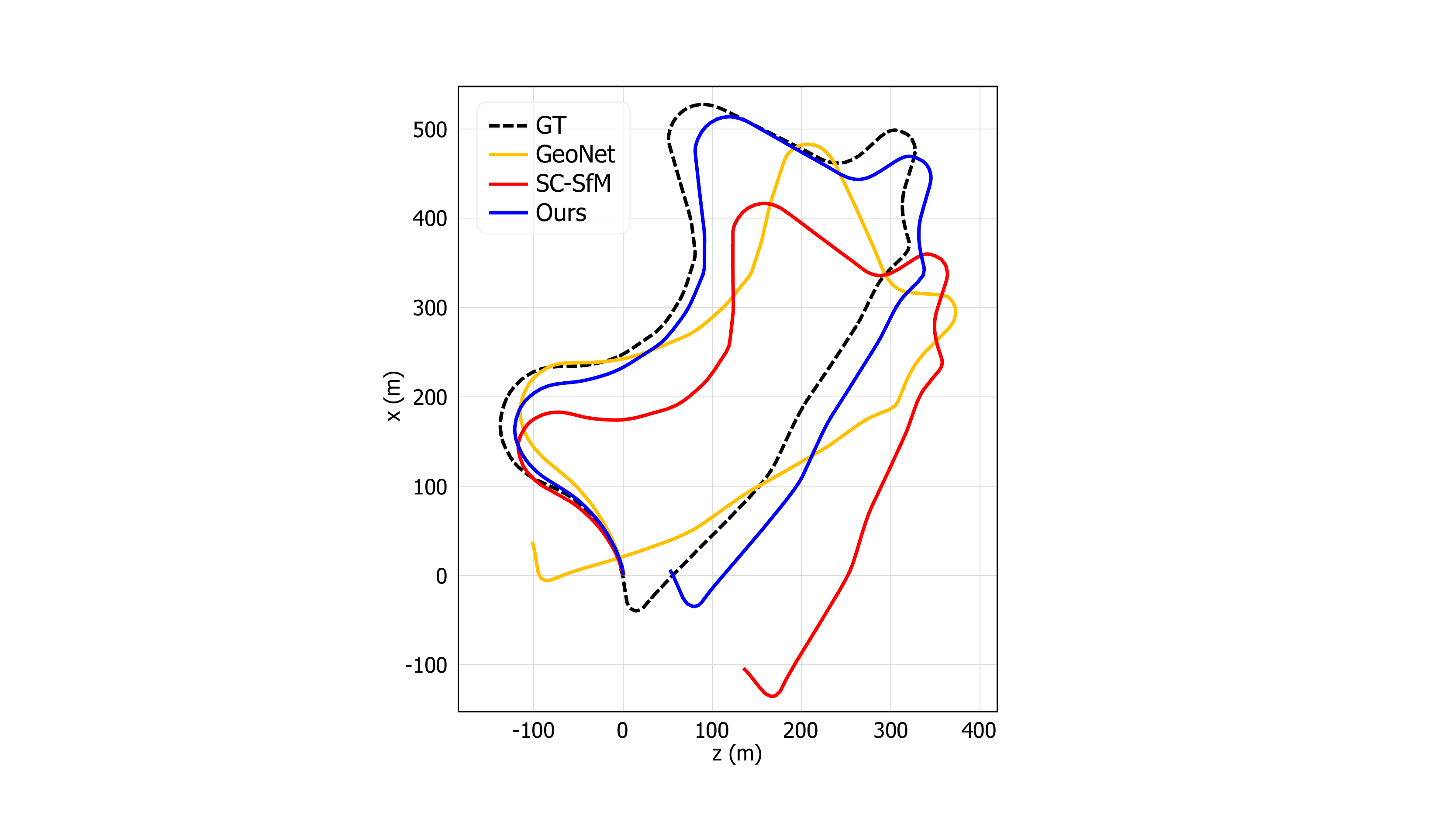}
	\vspace{-4mm}
	\captionof{figure}{Results on seq. 09.}
	\vspace{-8mm}
	\label{vo}
\end{minipage}
\vspace{-5mm}
\end{table}

\vspace{-2mm}
\section{Conclusions}
\vspace{-3mm}

In this work, we proposed a novel framework that predicts 6-DoF motion of multiple dynamic objects, ego-motion and depth with monocular image sequences.
Leveraging video instance segmentation, we design an unsupervised end-to-end joint training pipeline.
There are three main contributions of our work: (1) differentiable forward image warping, (2) instance-wise view-synthesis and geometric consistency loss, and (3) auto-annotation scheme for video instance segmentation.
We show that our method outperforms the existing methods that estimate object motion, ego-motion and depth.
We also show that each proposed module plays a role in improving the performance of our framework.


%
%
\bibliographystyle{splncs04}
\bibliography{egbib}
\end{document}